\documentclass[10pt,twocolumn,letterpaper]{article}
\usepackage[pagenumbers]{cvpr} 

\usepackage[utf8]{inputenc} 
\usepackage[T1]{fontenc}    
\usepackage{url}            
\usepackage{booktabs}       
\usepackage{amsfonts}       
\usepackage{nicefrac}       
\usepackage{microtype}      

\usepackage{graphicx}
\usepackage{amsmath}
\usepackage{amssymb}

\usepackage{diagbox}
\usepackage{multicol}
\usepackage{enumerate}
\usepackage{times}
\usepackage{epsfig}
\usepackage{threeparttable}
\usepackage{enumitem}
\usepackage{multirow}
\usepackage{color}
\usepackage{array}
\usepackage{setspace}
\usepackage{makecell}
\usepackage{indentfirst}
\usepackage{soul}

\def\OurModel{VLA-4D}

\newcommand{\gh}[1]{{\color{black}#1}}

\definecolor{cvprblue}{rgb}{0.21,0.49,0.74}
\usepackage[pagebackref,breaklinks,colorlinks,allcolors=cvprblue]{hyperref}

\title{VLA-4D: Embedding 4D Awareness into Vision-Language-Action Models \\ for SpatioTemporally Coherent Robotic Manipulation}

\author{Hanyu Zhou\textsuperscript{\rm 1}, Chuanhao Ma\textsuperscript{\rm 2}, Gim Hee Lee\textsuperscript{\rm 1}\\
  \textsuperscript{\rm 1} School of Computing, National University of Singapore\\
  \textsuperscript{\rm 2} School of Artificial Intelligence and Automation, Huazhong University of Science and Technology\\
  {\tt\small {\{hy.zhou, gimhee.lee\}}@nus.edu.sg}
 }

\usepackage[sort&compress]{natbib}
\begin{document}

\twocolumn[{
    \maketitle
    \setlength{\abovecaptionskip}{6pt}
      \setlength{\belowcaptionskip}{14pt}
      \centering
    \includegraphics[width=0.99\linewidth]{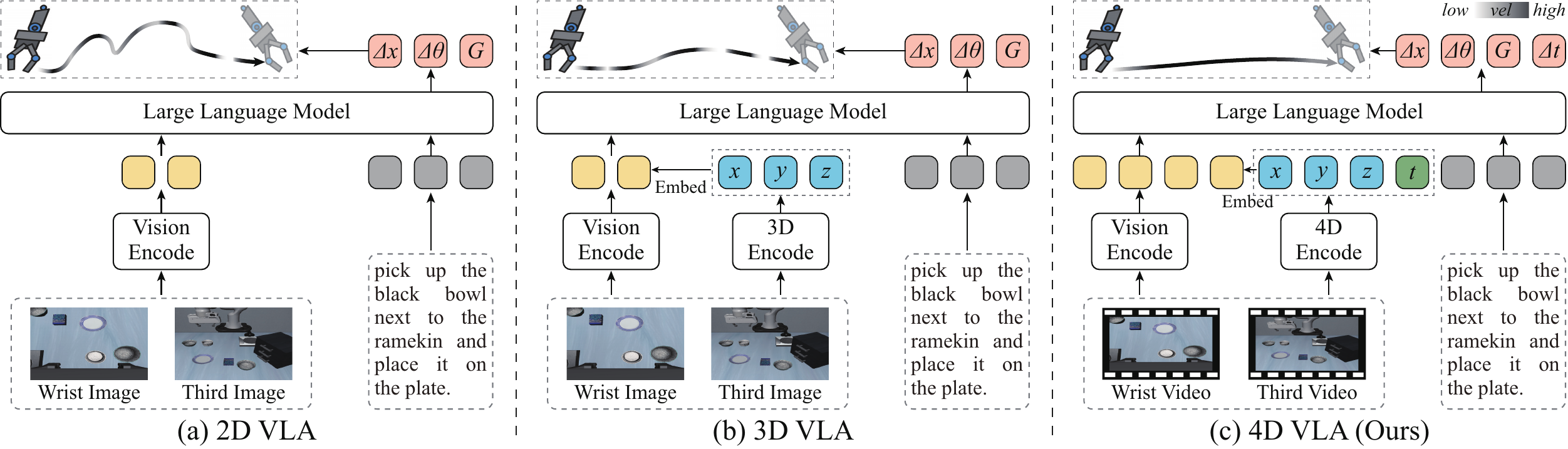}
    \captionof{figure}{Illustration of various VLA paradigms for robotic manipulation. (a) 2D VLAs encode visual and language modalities into the LLM to predict actions, which remain spatiotemporally discontinuous. (b) 3D VLAs embed 3D positions into visual representations to improve spatial precision and smoothness of actions, but lack temporal coherence. (c) Our VLA-4D integrates both 3D positions and 1D time into visual representations, and extends the action representation into the spatiotemporal domain to achieve spatiotemporal coherence. 
    }
    \label{Fig:Paradigm}
}]

\maketitle
\begin{abstract}
Vision-language-action (VLA) models show potential for general robotic tasks, but remain challenging in spatiotemporally coherent manipulation, which requires fine-grained representations. Typically, existing methods embed 3D positions into visual representations to enhance the spatial precision of actions. However, these methods struggle to achieve temporally coherent control over action execution. 
In this work, we propose \textbf{VLA-4D}, a general VLA model with 4D awareness for spatiotemporally coherent robotic manipulation. Our model is guided by two key designs: 1) \textbf{4D-aware visual representation}. We extract visual features, embed 1D time into 3D positions for 4D embeddings, and fuse them into a unified visual representation via a cross-attention mechanism. 2) \textbf{Spatiotemporal action representation.} We extend conventional spatial action representations with temporal information to enable the spatiotemporal planning, and align the multimodal representations into the LLM for spatiotemporal action prediction. Within this unified framework, the designed visual and action representations jointly make robotic manipulation spatially-smooth and temporally-coherent. In addition, we extend the VLA dataset with temporal action annotations for fine-tuning our model. Extensive experiments have been conducted to verify the superiority of our method across different tasks of robotic manipulation.
\end{abstract}

\section{Introduction}
\label{sec:intro}

\gh{With the success of vision-language models (VLMs) in scene reasoning~\cite{liu2023visual, chen2023pali, alayrac2022flamingo}, recent research explores their use in downstream robotics for manipulation~\cite{brohan2022rt, kim2024openvla, kim2025fine} and navigation~\cite{liu2024volumetric, wang2024vision}. As shown in \cref{Fig:Paradigm} (a), this has led to a specialized paradigm of vision-language-action (VLA) models that connect visual reasoning to action planning. Despite strong results on general robotic tasks~\cite{kim2024openvla, black2024pi_0, zitkovich2023rt, chi2025diffusion}, VLAs still struggle with spatiotemporally coherent manipulation that demands fine-grained representations. The key issues are: single-image inputs induce coarse visual reasoning, and 2D–3D (image-robot) coordinate system mismatches degrade action precision. Therefore, our goal is to strengthen fine-grained representations for visual reasoning and action planning.
}

\gh{
Most existing VLA models~\cite{zheng2024tracevla, qu2025spatialvla, zhen20243d} embed 3D positional cues into visual features to strengthen spatial reasoning, which improves action precision and smoothness as shown in \cref{Fig:Paradigm} (b). However, action planning is inherently spatiotemporal and demands continuity over time. Consequently, these 3D VLAs often struggle with fine-grained temporal control that leads to behaviors such as idle pauses or jitter. Recent works~\cite{niu2025pre, zhang20254d} add temporal signals such as frame indices and fuse them with 3D position embeddings in the visual stream. Although this helps temporal reasoning such as state precedence, it does not directly enforce temporally coherent action plans. We argue that coherent manipulation requires jointly enhancing spatiotemporal perception in both the visual and action representations within a VLA.
}


To address these issues, we propose to embed distinct spatiotemporal information into both the visual representation for reasoning and the action representation for planning. For 
visual representation, we 
\gh{suggest} that 3D positional information enhances 
understanding of scene geometry and spatial localization of subsequent actions, 
\gh{and} 1D temporal information further improves the perception of dynamic patterns and temporal states of actions. This indicates that 4D spatiotemporal cues play a crucial role in promoting both visual reasoning and action planning. For 
action representation, we observe that spatial control serves as the foundation for fine-grained planning, 
\gh{and} temporal control is indispensable for achieving a coherent action process. This motivates us to augment conventional spatial actions with temporal information 
\gh{for} coherent spatiotemporal operations. Consequently, these enhanced visual and action representations jointly unlock the potential of VLA models for spatiotemporally coherent robotic manipulation.

In this work, we propose \textbf{\OurModel}, a general vision-language-action model with 4D awareness for spatiotemporally coherent robotic manipulation. As illustrated in \cref{Fig:Paradigm} (c), our VLA-4D relies on a 4D-aware visual representation and a spatiotemporal action representation.
For the 4D-aware visual representation, we first encode the input video sequence into visual features and geometric features. 
Within the geometric space, we encode 3D positions and 1D time into 4D spatiotemporal embeddings. We then fuse the 4D embeddings into the visual features via a cross-attention mechanism.
For the spatiotemporal action representation, we extend the conventional spatial control parameters by introducing additional varying temporal variables
\gh{to enable} fine-grained spatiotemporal planning of robotic actions. By performing multimodal alignment, the LLM produces
spatiotemporal actions for the robot.
Under this unified framework, the 4D-aware visual representation and the spatiotemporal action representation jointly ensure the spatial smoothness and temporal coherence for robotic manipulation. In addition, we augment the LIBERO \cite{liu2023libero} dataset with temporal action annotations to fine-tune our model for more effective spatiotemporal robotic manipulation. Our main contributions are summarized as follows:
\begin{itemize}[leftmargin=10pt]
\item We propose \textbf{\OurModel}, a general 4D vision-language-action model for spatiotemporally coherent robotic manipulation. Our model embeds spatiotemporal information into both visual and action representations 
\gh{for} fine-grained visual reasoning and action planning.

\item We design an explicit 4D-aware visual representation. A cross-attention mechanism fuses 3D positions and 1D time into visual features for stronger fine-grained spatiotemporal reasoning of our model.

\item We construct a spatiotemporal action representation. Temporal control variables are incorporated into conventional spatial actions 
\gh{to improve} the spatial smoothness and temporal coherence of robotic operations.

\item We extend the VLA dataset with temporal action annotations to fine-tune our model. Extensive experiments demonstrate that our framework achieves state-of-the-art performance across various robotic manipulation tasks.
\end{itemize}

\section{Related Work}
\label{sec:related}
\noindent
\textbf{Vision-Language-Action Models.}
Vision-language models \cite{liu2023visual, alayrac2022flamingo, liu2023improved, li2023blip2, zhou2025llava} have achieved remarkable success in scene reasoning, inspiring a new class of end-to-end models for embodied intelligence, \ie, vision-language-action models. 
\gh{VLAs~\cite{kim2024openvla, team2024octo, zhou2025autovla, li2025simplevla, black2024pi_0, zhao2025cot, wen2025tinyvla} often encode actions as sequences of discrete, text-like tokens. These tokens are co-embedded with visual and linguistic inputs in a large language model~\cite{touvron2023llama, touvron2023llama2} for a motion-planning paradigm grounded in scene reasoning for robotic manipulation.
}
Although these VLA models indeed show great potential for general robotic tasks, they still face challenges in achieving spatiotemporally coherent manipulation. 
\gh{Robotic operations demand precise and continuous action parameters. In contrast, the knowledge learned directly from images and text is often coarse-grained. This mismatch makes it difficult to plan motions that are both accurate and temporally coherent.}
In this work, we focus on enhancing the fine-grained representation for visual reasoning and action planning.

\begin{figure*}
  \setlength{\abovecaptionskip}{5pt}
  \setlength{\belowcaptionskip}{-5pt}
  \centering
   \includegraphics[width=0.99\linewidth]{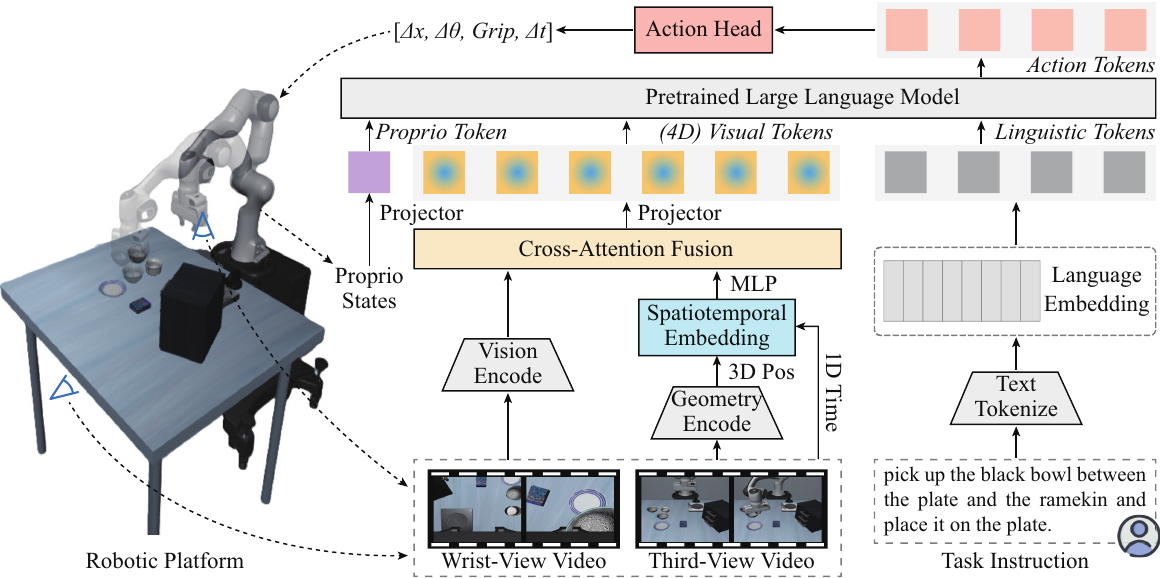}
   \caption{Our VLA-4D consists of two stages: 1) \textbf{4D-aware visual representation.} Encode 3D positions and 1D time into 4D spatiotemporal embeddings, and fuse them into visual features via a cross-attention mechanism. 2) \textbf{Spatiotemporal action representation.} Extend action parameters into the spatiotemporal domain, and align multimodal representations into the LLM for robotic action prediction.
   }
   \label{Fig:Framework}
\end{figure*}

\vspace{1mm}
\noindent
\textbf{3D Vision-Language-Action Models.}
For VLA models, the key to improving the action precision lies in 
\gh{improving} the fine-grained visual reasoning. Typically, existing VLA models \cite{qu2025spatialvla, zhen20243d, zheng2024tracevla, li20253ds, huang2025graphcot} embed 3D positional information into visual representations to strengthen spatial reasoning, thus optimizing the spatial control parameters of actions. Although these methods may achieve a relatively optimal spatial trajectory, they are still limited in achieving fine-grained reasoning and control along the temporal dimension, such as sequential state chaos and potential operational idling. 
\gh{We thus propose to embed 3D positions and 1D time into the visual representation to strengthen spatiotemporal reasoning and improve action precision in robotic manipulation.}

\vspace{1mm}
\noindent
\textbf{4D Vision-Language-Action Models.}
The research line most closely related to our work is 4D VLA models \cite{zhang20254d, niu2025pre, hou20254d}. These 4D models integrate spatial positional and temporal information into visual representations to further improve the prediction precision of robotic actions. For example, Niu \emph{et al.} \cite{niu2025pre} learn 4D point trajectories from videos to model visual representations and transfer the learned distribution to robotic control 
\gh{for} action prediction via an autoregressive model. 
\gh{Zhang \emph{et al}. \cite{zhang20254d} encode frame indices from past videos and fuse them with 3D positional embeddings in the visual stream. This mitigates temporal state ambiguity during robotic execution.}
Although these methods strengthen fine-grained spatiotemporal visual reasoning, they still struggle to explicitly improve the precision of temporal action planning and the overall coherence of operation execution. In contrast, we posit that it is essential to 
enhance the 4D awareness of visual representations, 
and to represent the spatiotemporal control parameters of robotic actions 
\gh{for} spatiotemporally coherent robotic manipulation.

\section{Our VLA-4D}
\label{sec:method}
\noindent
\textbf{Overview.}
\cref{Fig:Framework} shows the architecture of our VLA-4D with two key stages:
1) \emph{\textbf{4D-Aware Visual Representation,}} 
where visual features are enhanced by 4D spatiotemporal embeddings with cross-attention fusion.
2) \emph{\textbf{SpatioTemporal Action Representation,}} 
where action parameters are extended to the spatiotemporal domain and aligned with multimodal representations for task optimization. These components enable VLAs to achieve finer-grained spatiotemporal visual reasoning and action planning for robotic manipulation.

\vspace{2mm}
\noindent
\textbf{Our Framework.}
Given a video sequence $I$ and instruction texts, we employ a vision encoder to extract visual features $f_v$ and a text tokenizer to extract linguistic tokens $\tau_l$.

\vspace{1mm}
\noindent \textbf{1) \textit{4D-Aware Visual Representation (cf. \cref{sec:visualrepresentation})}.}
This stage enhances 4D perception of visual reasoning. We first encode the video sequence into a geometric latent space to obtain 3D positions $p_{3D}$. We then encode 1D time $t$ and 3D positions $p_{3D}$ into 4D learnable embeddings $f_{4D}$, which are fused with visual features through a cross-attention mechanism:
\begin{equation}
\setlength\abovedisplayskip{5pt}
\setlength\belowdisplayskip{5pt}
    \begin{aligned}
        f_{4D} = \operatorname{STE}(p_{3D}, t), \quad f_v^{4D} = \operatorname{CAtt}(f_v, f_{4D}),
    \label{eq:attention}
    \end{aligned}
\end{equation}
where $\operatorname{STE}$ denotes the spatiotemporal embedding operation. The unified visual representation $f_v^{4D}$ can perceive the 4D semantic and geometry of the scene for visual reasoning.

\vspace{1mm}
\noindent \textbf{2) \textit{SpatioTemporal Action Representation (cf. \cref{sec:actionrepresentation})}.}
This stage extends action planning into the spatiotemporal dimension.
We formulate the robotic action representation along the spatiotemporal dimension: $A = [X, T]$, where $X$, $T$ denote spatiotemporal control variables. The fused visual features $f_v^{4D}$ from the previous stage and the proprioceptive states $f_{p}$ are projected into language embedding space: $[\tau_v^{4D}, \tau_p] = \operatorname{Proj}(f_v^{4D}, f_{p})$, where $\tau_v^{4D}$, $\tau_p$ are the visual and proprioceptive tokens aligned with linguistic tokens $\tau_l$. The LLM with action head then learns the mapping from the aligned multimodal tokens to actions: $A=\mathcal{LLM}(\tau_v^{4D}, \tau_p, \tau_l)$.

\vspace{2mm}
\noindent
\textbf{Remarks.}
The output $A$ denotes the fine-grained spatiotemporal action parameters. Our unified framework enhances the 4D awareness of visual reasoning and enables spatiotemporally coherent action planning for robotic manipulation. The following sections detail the
design of each stage.

\subsection{4D-Aware Visual Representation}
\label{sec:visualrepresentation}
Existing VLA models utilize images and instructions to directly learn visual reasoning to enable action planning for robotic manipulation. However, the visual knowledge remains relatively coarse-grained and carries the risk of reducing the precision of the predicted action control. 
In this section, we explore how to enhance fine-grained visual representations to achieve precise robotic action planning.

\begin{figure}
  \setlength{\abovecaptionskip}{5pt}
  \setlength{\belowcaptionskip}{-5pt}
  \centering
   \includegraphics[width=0.99\linewidth]{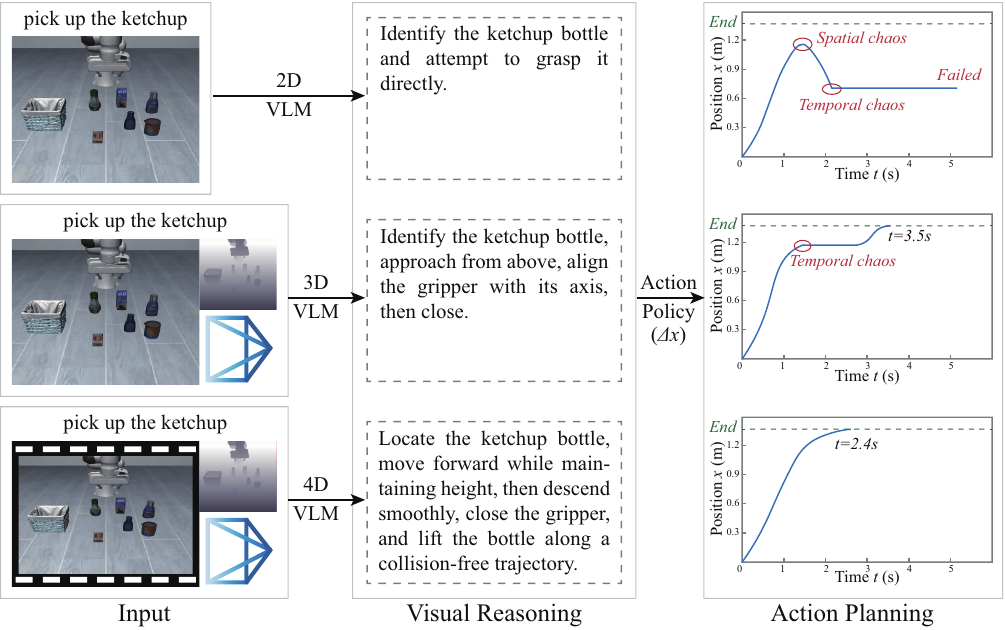}
   \caption{Effect of different visual representations. 3D spatial information enhances the understanding of scene geometry and subsequent action localization, while 1D temporal information further ensures the dynamic perception and temporal action state.
   }
   \label{Fig:4DVision}
\end{figure}

\vspace{1mm}
\noindent
\textbf{4D SpatioTemporal Embedding.}
\gh{Visual and action representations within VLAs exhibit certain discrepancies in the spatial and temporal dimensions. There is a coordinate system mismatch in the spatial dimension, where the image is in a 2D coordinate system and the action is in a 3D world (or robot) coordinate system. Furthermore, there is also a disagreement in the temporal dimension, where the image is a static snapshot of a scene and the action forms a sequence of 3D trajectories over time. The discrepancies in the spatial-temporal domains led us to consider the unification of the coordinate systems with temporal cues.}
To this end, we analyze the effects of visual inputs corresponding to different coordinate systems (2D pixel space \emph{v.s.} 3D world space) and temporal settings on visual reasoning and action planning. 
As shown in \cref{Fig:4DVision}, 3D coordinate information enhances the understanding of scene geometry and spatial localization of subsequent actions, while 1D temporal information promotes the dynamic perception and temporal action state.
Inspired by this \gh{observation}, we 
\gh{conjecture} that visual reasoning endowed with 4D awareness can facilitate fine-grained action planning. Given third-view and wrist-view video sequences, we adopt VGGT \cite{wang2025vggt} as the geometry encoder to extract the camera pose $P$ and depth $D$ at a timestamp $t$. Combined with intrinsic parameter $K$, we transform 2D pixel coordinate $p_{2D}$ to world (or robot) coordinate system via geometric projection \cite{zou2018df, zhou2017unsupervised}:
\begin{equation}
\setlength\abovedisplayskip{4pt}
\setlength\belowdisplayskip{4pt}
    \begin{aligned}
        p_{3D} = P^{-1}(DK^{-1}p_{2D}).
    \label{eq:position}
    \end{aligned}
\end{equation}

After traversing all timestamps for geometric extraction, we propose a spatiotemporal embedding operation $\operatorname{STE}$ to integrate 3D positions and 1D time. Specifically, we introduce a Fourier-based encoding strategy \cite{li2021learnable} to convert positions and timestamps into learnable patterns, and map them into a 4D representation through a linear layer:
\begin{equation}
\setlength\abovedisplayskip{4pt}
\setlength\belowdisplayskip{4pt}
    \begin{aligned}
        \psi(x)=&1/\sqrt{d}~[cos(xW^{\top}_r)~||~sin(xW^{\top}_r)],\\
        &f_{4D} = w_p\cdot[\psi(p_{3D})~||~\psi(t)],
    \label{eq:4drepresentation}
    \end{aligned}
\end{equation}
where $d$ denotes the dimension and $W_r$ is the learnable parameter of the Fourier feature. The resulting $f_{4D}$ represents the 4D spatiotemporal geometry of the scene.

\vspace{1mm}
\noindent
\textbf{Cross-Attention Fusion.}
In our model, we employ a ViT variant \cite{bai2025qwen2} as the vision encoder to extract high-level visual features with semantic information. However, such standalone features cannot be localized to the world coordinate system or aligned with the temporal sequence state required for robotic manipulation. This motivates us to further embed 4D awareness into the visual features. We first introduce a Multi-Layer Perceptron (MLP) to make the dimension of the 4D spatiotemporal embeddings the same as the dimension of the visual features: $\hat{f}_{4D} = \operatorname{MLP}(f_{4D})$. Next, we fuse the 4D spatiotemporal embeddings into the visual features via a cross-attention mechanism \cite{vaswani2017attention, chen2021crossvit}:
\begin{equation}
\setlength\abovedisplayskip{4pt}
\setlength\belowdisplayskip{4pt}
    \begin{aligned}
        q&=w_qf_v, ~k=w_k\hat{f}_{4D}, ~v=w_v\hat{f}_{4D},\\
        &f_v^{4D} = f_v + \operatorname{softmax}(qk^{\top}\sqrt{d})v,
    \label{eq:cross_attention}
    \end{aligned}
\end{equation}
where $w$ is a learnable weight. The fused result $f_v^{4D}$ denotes the unified visual representation that can capture both the 4D semantic and geometric characteristics of the scene for spatiotemporal visual reasoning.

\subsection{SpatioTemporal Action Representation}
\label{sec:actionrepresentation}
Ideally, robotic motion planning should be a spatially smooth and temporally coherent dynamic process.
Although 4D-aware visual representations facilitate fine-grained action prediction, they are not sufficient to explicitly ensure the overall coherence of robotic operations. Therefore, our goal is to build the fine-grained spatiotemporal action representation for coherent robotic manipulation.

\begin{figure}
  \setlength{\abovecaptionskip}{7pt}
  \setlength{\belowcaptionskip}{-5pt}
  \centering
   \includegraphics[width=0.99\linewidth]{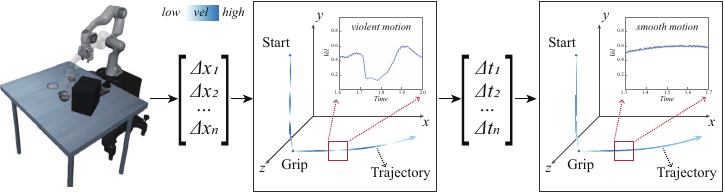}
   \caption{Illustration of spatiotemporal action representation. Spatial parameters enable fine-grained action planning, while temporal parameters further improve the action coherence during execution.
   }
   \label{Fig:STAction}
\end{figure}

\vspace{1mm}
\noindent
\textbf{Spatiotemporal Action Definition.}
Conventional action representations \cite{kim2024openvla, chi2025diffusion} 
\gh{focus mainly} on spatial control parameters 
\gh{such as translation} for fine-grained robotic operations. 
\gh{However, such representations ignore the specific execution duration and enable the robot to learn only spatial movements within an action step. 
This omission increases the risk of premature termination or delayed responses in robotic tasks, which leads to a discontinuous control process.}
As illustrated in \cref{Fig:STAction}, we analyze the effects of spatial and temporal control parameters on the robotic operation process. Note that the variable time here is predefined according to the parameters of the robot and task configurations. We can observe that spatial actions form the foundation of fine-grained robotic operations, 
\gh{and the incorporation of} reasonable temporal parameters further promotes the continuity and coherence of the overall action execution process.
Driven by this motivation, we augment conventional spatial action representations with temporal information. Specifically, the conventional spatial action is defined as $X=[\Delta x, \Delta \theta, Grip]$, where $\Delta x$ denotes the translational displacement of the end-effector, $\Delta \theta$ represents its rotational change in orientation, and $Grip$ specifies the gripper control signal indicating the open or close state.
We further formulate a temporal action representation: $T=\Delta t$, which is a time variable for step-level action control jointly determined by the scene content \emph{corresponding to vision in VLA}, the operation task \emph{corresponding to language in VLA}, and the robotic embodiment \emph{corresponding to proprioceptive state feedback}. In this way, we extend the spatial action into a spatiotemporal action: $A=[\Delta x, \Delta \theta, Grip, \Delta t]$.

\vspace{1mm}
\noindent
\textbf{Multimodal Alignment and Optimization.}
Similar to conventional VLMs \cite{liu2023visual, li2022blip, zhou2025uni4d, zhou2025llafea}, our VLA model produces spatiotemporal actions through two key components: multimodal representation alignment and task optimization. For 
multimodal alignment, we align 
vision, language, and additional proprioceptive states into a unified learnable space.
Considering that the input to a large language model requires text-like tokens, we first introduce a multi-layer perceptron as the projection function $\operatorname{Proj}(\cdot)$ to map the 4D-aware visual features and proprioceptive states into the language embedding space: $\tau_v^{4D}=\operatorname{MLP}(f_v^{4D})$, $\tau_p=\operatorname{MLP}(f_p)$, where $\tau_v^{4D}$ and $\tau_p$ denote the visual and proprioceptive tokens. Next, we tokenize the input instruction into the same language embedding space to obtain the linguistic tokens $\tau_l$.

For 
task optimization, 
\gh{our goal is} to learn a model that maps the aligned multimodal representations to spatiotemporal actions. We first introduce a pretrained large language model $\mathcal{T}(\cdot)$ and append an MLP-based action head $\mathcal{H}(\cdot)$. We then concatenate the visual, linguistic, and proprioceptive tokens, and feed them into the model for action prediction: 
\begin{equation}
\setlength\abovedisplayskip{4pt}
\setlength\belowdisplayskip{4pt}
    \begin{aligned}
        [\Delta x,~ \Delta \theta, ~Grip, ~\Delta t] = \mathcal{H}(\mathcal{T}([\tau_v^{4D}, ~\tau_p, ~\tau_l])).
    \label{eq:actionrepresentation}
    \end{aligned}
\end{equation}

To train the model for precise spatiotemporal action prediction, we further apply an L1-norm loss \cite{huber1992robust} on the predicted spatiotemporal action variables: 
\begin{equation}
\setlength\abovedisplayskip{4pt}
\setlength\belowdisplayskip{4pt}
    \begin{aligned}
        \mathcal{L}_{action}&=\sum (|\Delta x - \tilde{\Delta x}|_1
        + |\Delta \theta - \tilde{\Delta \theta}|_1 \\
        &+ |Grip - \tilde{Grip}|_1
        + |\Delta t - \tilde{\Delta t}|_1),
    \label{eq:loss}
    \end{aligned}
\end{equation}
where variables with a tilde (~$\tilde{\cdot}$~) denote ground-truth action values. During inference, our VLA-4D can directly predict future spatiotemporal actions based on historical videos, proprioceptive states, and task-specific textual instructions 
\gh{for} fine-grained and coherent robotic manipulation.

\begin{table*}\footnotesize
    \setlength{\abovecaptionskip}{3pt}
    \setlength\tabcolsep{1.2pt}
    \setlength{\belowcaptionskip}{3pt}
    \caption{Quantitative results of VLAs for fine-tuned robotic manipulation tasks on the LIBERO benchmark.} 
  \centering
  \renewcommand\arraystretch{1.28}
  \begin{tabular}{cccccccccccc}
    \Xhline{1pt}
      \multicolumn{2}{c|}{\multirow{2}{*}{Methods}} &
      \multicolumn{2}{c|}{\multirow{1}{*}{Spatial}} &
      \multicolumn{2}{c|}{\multirow{1}{*}{Object}} &
      \multicolumn{2}{c|}{\multirow{1}{*}{Goal}} &
      \multicolumn{2}{c|}{\multirow{1}{*}{Long}} &
      \multicolumn{2}{c}{\multirow{1}{*}{Average}} \\      
      \cline{3-12}
      \multicolumn{2}{c|}{\multirow{1}{*}{}} &
      \multicolumn{1}{c}{\multirow{1}{*}{Succ. rate(\%)$\uparrow$}} &
      \multicolumn{1}{c|}{\multirow{1}{*}{Time(s)$\downarrow$}} &
      \multicolumn{1}{c}{\multirow{1}{*}{Succ. rate(\%)$\uparrow$}} &
      \multicolumn{1}{c|}{\multirow{1}{*}{Time(s)$\downarrow$}} &
      \multicolumn{1}{c}{\multirow{1}{*}{Succ. rate(\%)$\uparrow$}} &
      \multicolumn{1}{c|}{\multirow{1}{*}{Time(s)$\downarrow$}} &
      \multicolumn{1}{c}{\multirow{1}{*}{Succ. rate(\%)$\uparrow$}} &
      \multicolumn{1}{c|}{\multirow{1}{*}{Time(s)$\downarrow$}} &
      \multicolumn{1}{c}{\multirow{1}{*}{Succ. rate(\%)$\uparrow$}} &
      \multicolumn{1}{c}{\multirow{1}{*}{Time(s)$\downarrow$}} \\

      \hline 
      \multicolumn{1}{c|}{\multirow{4}{*}{2D}} &
      \multicolumn{1}{c|}{\multirow{1}{*}{OpenVLA \cite{kim2024openvla}}} &
      \multicolumn{1}{c}{\multirow{1}{*}{84.7 $\pm$ 0.9}} &
      \multicolumn{1}{c|}{\multirow{1}{*}{5.5}} &
      \multicolumn{1}{c}{\multirow{1}{*}{88.4 $\pm$ 0.8}} &
      \multicolumn{1}{c|}{\multirow{1}{*}{7.5}} &
      \multicolumn{1}{c}{\multirow{1}{*}{79.2 $\pm$ 1.0}} &
      \multicolumn{1}{c|}{\multirow{1}{*}{6.1}} &
      \multicolumn{1}{c}{\multirow{1}{*}{53.7 $\pm$ 1.3}} &
      \multicolumn{1}{c|}{\multirow{1}{*}{13.1}} &
      \multicolumn{1}{c}{\multirow{1}{*}{76.5 $\pm$ 0.6}} &
      \multicolumn{1}{c}{\multirow{1}{*}{8.1}} \\

      \multicolumn{1}{c|}{\multirow{1}{*}{}} &
      \multicolumn{1}{c|}{\multirow{1}{*}{Octo \cite{team2024octo}}} &
      \multicolumn{1}{c}{\multirow{1}{*}{78.9 $\pm$ 1.0}} &
      \multicolumn{1}{c|}{\multirow{1}{*}{5.7}} &
      \multicolumn{1}{c}{\multirow{1}{*}{85.7 $\pm$ 0.9}} &
      \multicolumn{1}{c|}{\multirow{1}{*}{6.9}} &
      \multicolumn{1}{c}{\multirow{1}{*}{84.6 $\pm$ 0.9}} &
      \multicolumn{1}{c|}{\multirow{1}{*}{6.3}} &
      \multicolumn{1}{c}{\multirow{1}{*}{51.1 $\pm$ 1.3}} &
      \multicolumn{1}{c|}{\multirow{1}{*}{9.3}} &
      \multicolumn{1}{c}{\multirow{1}{*}{75.1 $\pm$ 0.6}} &
      \multicolumn{1}{c}{\multirow{1}{*}{7.1}} \\

      \multicolumn{1}{c|}{\multirow{1}{*}{}} &
      \multicolumn{1}{c|}{\multirow{1}{*}{DiffusionPolicy \cite{chi2025diffusion}}} &
      \multicolumn{1}{c}{\multirow{1}{*}{78.3 $\pm$ 1.1}} &
      \multicolumn{1}{c|}{\multirow{1}{*}{6.4}} &
      \multicolumn{1}{c}{\multirow{1}{*}{92.5 $\pm$ 0.7}} &
      \multicolumn{1}{c|}{\multirow{1}{*}{7.8}} &
      \multicolumn{1}{c}{\multirow{1}{*}{68.3 $\pm$ 1.2}} &
      \multicolumn{1}{c|}{\multirow{1}{*}{6.4}} &
      \multicolumn{1}{c}{\multirow{1}{*}{50.5 $\pm$ 1.3}} &
      \multicolumn{1}{c|}{\multirow{1}{*}{15.2}} &
      \multicolumn{1}{c}{\multirow{1}{*}{72.4 $\pm$ 0.7}} &
      \multicolumn{1}{c}{\multirow{1}{*}{8.7}} \\

      \multicolumn{1}{c|}{\multirow{1}{*}{}} &
      \multicolumn{1}{c|}{\multirow{1}{*}{CogACT \cite{li2024cogact}}} &
      \multicolumn{1}{c}{\multirow{1}{*}{87.5 $\pm$ 0.9}} &
      \multicolumn{1}{c|}{\multirow{1}{*}{5.4}} &
      \multicolumn{1}{c}{\multirow{1}{*}{90.2 $\pm$ 1.1}} &
      \multicolumn{1}{c|}{\multirow{1}{*}{6.8}} &
      \multicolumn{1}{c}{\multirow{1}{*}{78.4 $\pm$ 0.8}} &
      \multicolumn{1}{c|}{\multirow{1}{*}{5.9}} &
      \multicolumn{1}{c}{\multirow{1}{*}{53.2 $\pm$ 1.2}} &
      \multicolumn{1}{c|}{\multirow{1}{*}{10.7}} &
      \multicolumn{1}{c}{\multirow{1}{*}{76.5 $\pm$ 0.9}} &
      \multicolumn{1}{c}{\multirow{1}{*}{7.0}} \\

      \hline 

      \multicolumn{1}{c|}{\multirow{2}{*}{3D}} &
      \multicolumn{1}{c|}{\multirow{1}{*}{TraceVLA \cite{zheng2024tracevla}}} &
      \multicolumn{1}{c}{\multirow{1}{*}{84.6 $\pm$ 0.2}} &
      \multicolumn{1}{c|}{\multirow{1}{*}{--}} &
      \multicolumn{1}{c}{\multirow{1}{*}{85.2 $\pm$ 0.4}} &
      \multicolumn{1}{c|}{\multirow{1}{*}{--}} &
      \multicolumn{1}{c}{\multirow{1}{*}{75.1 $\pm$ 0.3}} &
      \multicolumn{1}{c|}{\multirow{1}{*}{--}} &
      \multicolumn{1}{c}{\multirow{1}{*}{54.1 $\pm$ 1.0}} &
      \multicolumn{1}{c|}{\multirow{1}{*}{--}} &
      \multicolumn{1}{c}{\multirow{1}{*}{74.8 $\pm$ 0.4}} &
      \multicolumn{1}{c}{\multirow{1}{*}{--}} \\

      \multicolumn{1}{c|}{\multirow{1}{*}{}} &
      \multicolumn{1}{c|}{\multirow{1}{*}{SpatialVLA \cite{qu2025spatialvla}}} &
      \multicolumn{1}{c}{\multirow{1}{*}{88.2 $\pm$ 0.5}} &
      \multicolumn{1}{c|}{\multirow{1}{*}{5.3}} &
      \multicolumn{1}{c}{\multirow{1}{*}{89.9 $\pm$ 0.7}} &
      \multicolumn{1}{c|}{\multirow{1}{*}{6.4}} &
      \multicolumn{1}{c}{\multirow{1}{*}{78.6 $\pm$ 0.6}} &
      \multicolumn{1}{c|}{\multirow{1}{*}{5.9}} &
      \multicolumn{1}{c}{\multirow{1}{*}{55.5 $\pm$ 1.0}} &
      \multicolumn{1}{c|}{\multirow{1}{*}{8.9}} &
      \multicolumn{1}{c}{\multirow{1}{*}{78.1 $\pm$ 0.7}} &
      \multicolumn{1}{c}{\multirow{1}{*}{6.8}} \\

      \hline

      \multicolumn{1}{c|}{\multirow{2}{*}{4D}} &
      \multicolumn{1}{c|}{\multirow{1}{*}{4D-VLA \cite{zhang20254d}}} &
      \multicolumn{1}{c}{\multirow{1}{*}{88.9 $\pm$ 0.5}} &
      \multicolumn{1}{c|}{\multirow{1}{*}{--}} &
      \multicolumn{1}{c}{\multirow{1}{*}{95.2 $\pm$ 0.3}} &
      \multicolumn{1}{c|}{\multirow{1}{*}{--}} &
      \multicolumn{1}{c}{\multirow{1}{*}{90.9 $\pm$ 0.4}} &
      \multicolumn{1}{c|}{\multirow{1}{*}{--}} &
      \multicolumn{1}{c}{\multirow{1}{*}{79.1 $\pm$ 1.2}} &
      \multicolumn{1}{c|}{\multirow{1}{*}{--}} &
      \multicolumn{1}{c}{\multirow{1}{*}{88.6 $\pm$ 0.3}} &
      \multicolumn{1}{c}{\multirow{1}{*}{--}} \\

      \multicolumn{1}{c|}{\multirow{1}{*}{}} &
      \multicolumn{1}{c|}{\multirow{1}{*}{VLA-4D (Ours)}} &
      \multicolumn{1}{c}{\multirow{1}{*}{\textbf{97.9 $\pm$ 0.2}}} &
      \multicolumn{1}{c|}{\multirow{1}{*}{\textbf{4.1}}} &
      \multicolumn{1}{c}{\multirow{1}{*}{\textbf{98.6 $\pm$ 0.3}}} &
      \multicolumn{1}{c|}{\multirow{1}{*}{\textbf{5.6}}} &
      \multicolumn{1}{c}{\multirow{1}{*}{\textbf{97.8 $\pm$ 0.3}}} &
      \multicolumn{1}{c|}{\multirow{1}{*}{\textbf{4.6}}} &
      \multicolumn{1}{c}{\multirow{1}{*}{\textbf{94.8 $\pm$ 0.8}}} &
      \multicolumn{1}{c|}{\multirow{1}{*}{\textbf{6.9}}} &
      \multicolumn{1}{c}{\multirow{1}{*}{\textbf{97.4 $\pm$ 0.3}}} &
      \multicolumn{1}{c}{\multirow{1}{*}{\textbf{5.8}}} \\

       \Xhline{1pt}
       
  \end{tabular} 
   \label{Tab:Comparison}
\end{table*}

\begin{figure*}
  \setlength{\abovecaptionskip}{7pt}
  \setlength{\belowcaptionskip}{-5pt}
  \centering
   \includegraphics[width=0.99\linewidth]{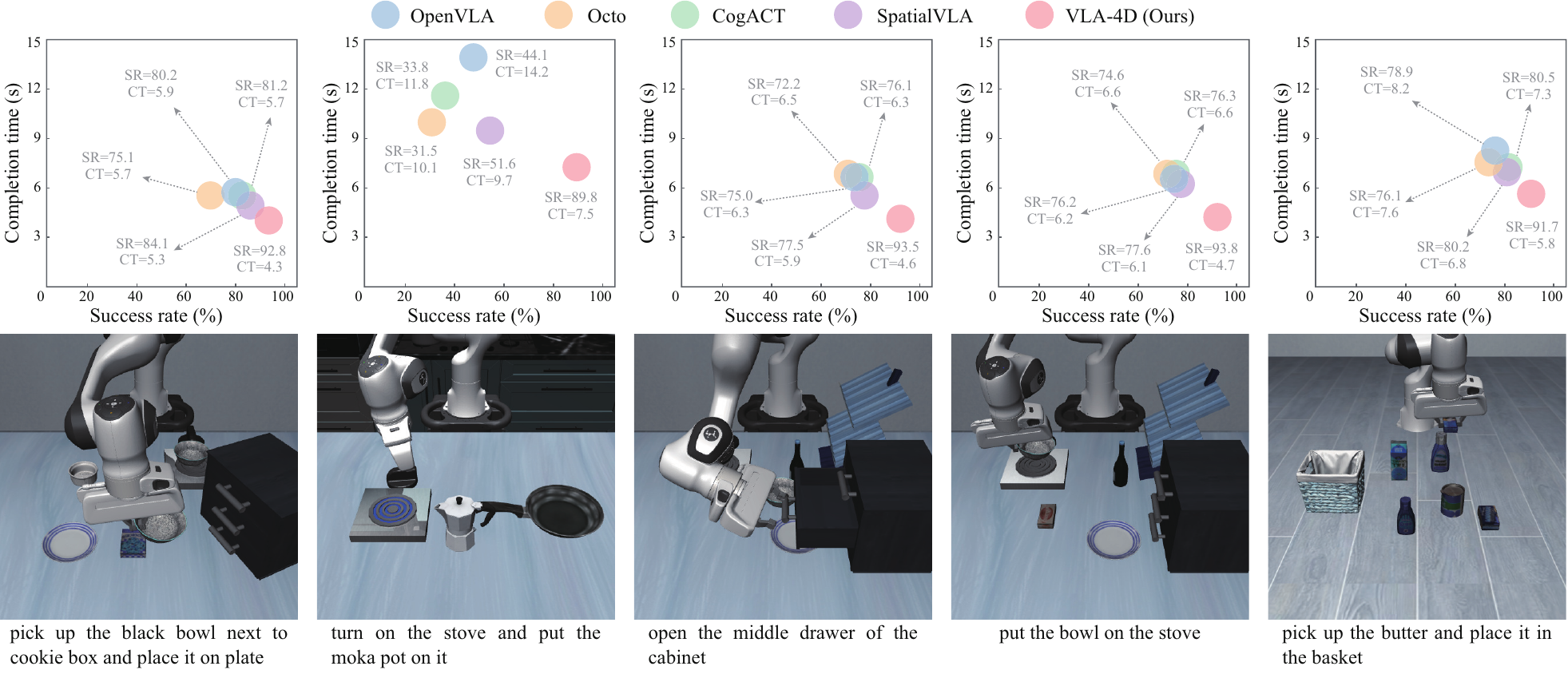}
   \caption{Quantitative comparison of VLAs on zero-shot robotic manipulation tasks.
   }
   \label{Fig:Comparison}
\end{figure*}

\section{Dataset and Training Pipeline}
\label{sec:training}
\subsection{Robotic Dataset}
Many robotic datasets \cite{o2024open, liu2023libero, khazatsky2024droid, zhao2023learning} have been proposed to evaluate the performance of VLAs. However, there is currently no dataset specifically designed for spatiotemporally coherent robotic manipulation in VLAs. To address this limitation, we select a representative robotic dataset \cite{liu2023libero} and extend its input modalities and action annotations to fine-tune our model for evaluating robotic tasks.

\vspace{1mm}
\noindent
\textbf{LIBERO.} 
LIBERO \cite{liu2023libero} is a simulation suite with four benchmark settings designed to advance lifelong learning in robotic manipulation, including spatial reasoning (LIBERO-Spatial), object understanding (LIBERO-Object), task goal (LIBERO-Goal) and long-horizon planning (LIBERO-Long). Note that 
LIBERO-Average denotes the whole dataset for task evaluation. For the input modalities, we render human-designed trajectories in a simulated environment at a fixed sampling frequency to obtain camera parameters, multi-view videos with timestamps, depths, and proprioceptive states. For the action annotations, without altering the original spatial action parameters, we manually select action chunks that exhibit a consistent motion trend and convert their step counts to variable temporal action annotations based on the sampling frequency. After data cleaning and selection, the final dataset contains 40 subtasks 
\gh{with a} total of 150k paired vision–language–action samples.

\subsection{Training Pipeline}
Our VLA-4D first loads the weights of several pretrained models \cite{bai2025qwen2, wang2025vggt} 
\gh{to initialize} its multimodal understanding capability, and then performs fine-tuning for robotic tasks. To fully enhance the learning capability of visual representations and improve the performance of action representations, we divide the training process into two stages:

\vspace{1mm}
\noindent
\textbf{Stage 1: 4D Vision-Language Alignment.}
This stage trains the 4D visual representations for reasoning within our VLM architecture. We select several 3D and 4D vision-language datasets 
\gh{including Scan2Cap \cite{chen2021scan2cap}, ScanQA \cite{azuma2022scanqa}, ScanRef \cite{chen2020scanrefer}, Multi3DRefer \cite{zhang2023multi3drefer}, Chat4D \cite{zhou2025llava}}
to train the VLM component of our model using only the LLM loss \cite{touvron2023llama, radford2018improving}. This ensures that the visual representations acquire strong 4D spatiotemporal perception and can interact with language. At this stage, we update the weights of cross-attention, spatiotemporal embedding, projector, and optimize the LLM component using LoRA adapters \cite{hu2022lora} while freezing the vision encoder, geometry encoder, and action head.

\vspace{1mm}
\noindent
\textbf{Stage 2: Robotic Task Fine-Tuning.}
This stage trains the spatiotemporal action representations for planning within our entire VLA architecture. We employ the modified LIBERO dataset with spatiotemporal action annotations as the training settings to fine-tune our model using the loss $\mathcal{L}_{action}$. This enables our VLA-4D to produce a series of spatiotemporal actions conditioned on task instructions and 4D vision inputs for diverse robotic manipulation tasks. At this stage, we update the weights of \gh{the} action head and projector, optimize cross-attention and LLM components using LoRA adapters \cite{hu2022lora} while freezing the remaining modules.

\begin{figure*}
  \setlength{\abovecaptionskip}{7pt}
  \setlength{\belowcaptionskip}{-5pt}
  \centering
   \includegraphics[width=0.99\linewidth]{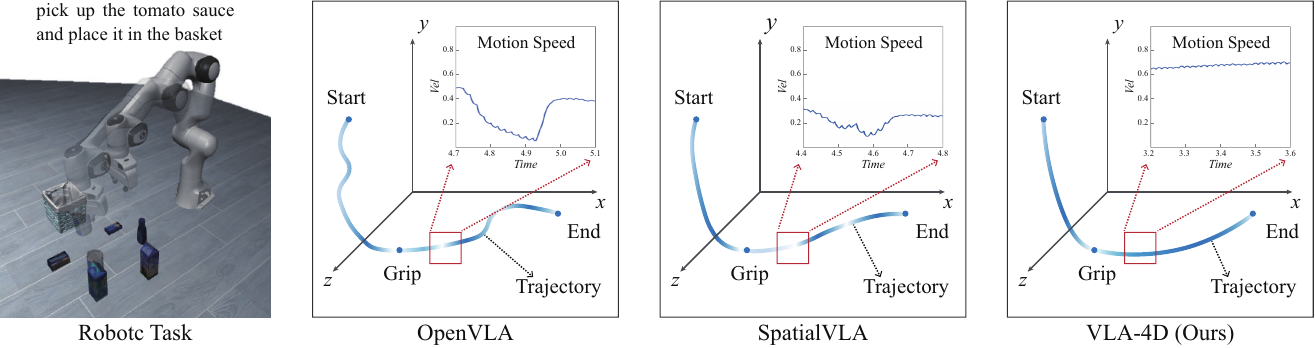}
   \caption{
   Visual comparison of VLAs on spatiotemporal action planning.
   } 
   \label{Fig:Comparison_Spatiotemporal}
\end{figure*}

\section{Experiments}
\label{sec:experiments}

\subsection{Experiment Setup}
\noindent
\textbf{Implementation Details.}
Our VLA-4D model utilizes the pre-trained weights of Qwen2.5-VL-7B \cite{bai2025qwen2} as the VLM (vision encoder + LLM) backbone, and VGGT \cite{wang2025vggt} as the geometry encoder. The cross-attention fusion module is a Transformer-based network architecture.
The whole model is trained on 8 RTX 6000 Ada GPUs using AdamW as the optimizer. In training stage 1, we set the learning rate to $1.0e-4$ with a batch size of 16. In training stage 2, we set the learning rate to $5.0e-5$ with a batch size of 24.

\vspace{1mm}
\noindent
\textbf{Comparison Methods.}
We compare our model with several typical VLAs: OpenVLA \cite{kim2024openvla}, Octo \cite{team2024octo}, CogACT \cite{li2024cogact}, DiffusionPolicy \cite{chi2025diffusion}, TraceVLA \cite{zheng2024tracevla}, SpatialVLA \cite{qu2025spatialvla}, \gh{and} 4D-VLA \cite{zhang20254d}. The first four methods are 2D VLA models using single images as inputs, the 
\gh{next} two methods are 3D VLA models using images and 3D positions as inputs, and the last 
method is a 4D VLA model using images, 3D positions, and 1D time as inputs.

\vspace{1mm}
\noindent
\textbf{Evaluation Metric.}
We compare all competing methods on the LIBERO dataset.
We set two types of evaluation settings. The first directly fine-tunes competing models and evaluates them across all robotic subtasks, 
\gh{and} the second evaluates the zero-shot generalization performance of competing models on a selected subset of subtasks. Moreover, we adopt task success rate (SR) and completion time (CT) as 
primary evaluation metrics.
Note that the experiments in \cref{sec:abalation} and \cref{sec:discussion} are evaluated on the LIBERO-Spatial and LIBERO-Goal benchmarks.

\subsection{Comparison with State-of-the-Art Models}
\label{sec:comparison}
\noindent
\textbf{Comparison on Fine-Tuned Tasks.}
In \cref{Tab:Comparison}, we compare the fine-tuning performance of competing methods on the LIBERO benchmark. We draw three key conclusions from the results. First, 2D models exhibit relatively inferior performance on complex real-world robotic manipulation tasks 
\gh{while} 3D and 4D models achieve significantly better results. Second, 4D models consistently outperform all other methods. 
\gh{The main reason is 4D models add 3D positions to align visual and action coordinate frames, and 1D time to resolve temporal ambiguity in planning. This joint space–time encoding stabilizes execution and reduces spurious or unstable actions.}
Third, our model achieves a higher task success rate than other competing models 
\gh{and requires} the shortest task completion time. Overall, our VLA-4D provides an effective paradigm for spatiotemporal robotic manipulation.

\begin{table}\footnotesize
    \setlength{\abovecaptionskip}{3pt}
    \setlength\tabcolsep{1.5pt}
    \setlength{\belowcaptionskip}{3pt}
    \caption{Effect of visual representation modules.} 
  \centering
  \renewcommand\arraystretch{1.3}
  \begin{tabular}{ccccccc}
    \Xhline{1pt}
      \multicolumn{1}{c}{\multirow{2}{*}{\makecell{Spatial\\embed}}} & \multicolumn{1}{c}{\multirow{2}{*}{\makecell{Temporal\\embed}}} & \multicolumn{1}{c|}{\multirow{2}{*}{\makecell{Feature\\fusion}}} &
      \multicolumn{2}{c|}{\multirow{1}{*}{LIBERO-Spatial}} &
      \multicolumn{2}{c}{\multirow{1}{*}{LIBERO-Goal}} \\

      \cline{4-5}\cline{6-7}
      \multicolumn{1}{c}{\multirow{1}{*}{}} &
      \multicolumn{1}{c}{\multirow{1}{*}{}} &
      \multicolumn{1}{c|}{\multirow{1}{*}{}} &
      \multicolumn{1}{c}{\multirow{1}{*}{Succ(\%)$\uparrow$}} &
      \multicolumn{1}{c|}{\multirow{1}{*}{Time(s)$\downarrow$}} &
      \multicolumn{1}{c}{\multirow{1}{*}{Succ(\%)$\uparrow$}} &
      \multicolumn{1}{c}{\multirow{1}{*}{Time(s)$\downarrow$}} \\

      \hline 
      \multicolumn{1}{c}{\multirow{1}{*}{$\times$}} &
      \multicolumn{1}{c}{\multirow{1}{*}{$\times$}} &
      \multicolumn{1}{c|}{\multirow{1}{*}{$\times$}} &
      \multicolumn{1}{c}{\multirow{1}{*}{89.4 $\pm$ 0.6}} &
      \multicolumn{1}{c|}{\multirow{1}{*}{5.7}} &
      \multicolumn{1}{c}{\multirow{1}{*}{90.1 $\pm$ 0.7}} &
      \multicolumn{1}{c}{\multirow{1}{*}{6.3}} \\

      \multicolumn{1}{c}{\multirow{1}{*}{$\surd$}} &
      \multicolumn{1}{c}{\multirow{1}{*}{$\times$}} &
      \multicolumn{1}{c|}{\multirow{1}{*}{$\times$}} &
      \multicolumn{1}{c}{\multirow{1}{*}{92.2 $\pm$ 0.4}} &
      \multicolumn{1}{c|}{\multirow{1}{*}{5.1}} &
      \multicolumn{1}{c}{\multirow{1}{*}{94.3 $\pm$ 0.5}} &
      \multicolumn{1}{c}{\multirow{1}{*}{5.6}} \\

      \multicolumn{1}{c}{\multirow{1}{*}{$\surd$}} &
      \multicolumn{1}{c}{\multirow{1}{*}{$\surd$}} &
      \multicolumn{1}{c|}{\multirow{1}{*}{$\times$}} &
      \multicolumn{1}{c}{\multirow{1}{*}{96.5 $\pm$ 0.3}} &
      \multicolumn{1}{c|}{\multirow{1}{*}{4.4}} &
      \multicolumn{1}{c}{\multirow{1}{*}{95.7 $\pm$ 0.4}} &
      \multicolumn{1}{c}{\multirow{1}{*}{4.9}} \\

      \multicolumn{1}{c}{\multirow{1}{*}{$\surd$}} &
      \multicolumn{1}{c}{\multirow{1}{*}{$\surd$}} &
      \multicolumn{1}{c|}{\multirow{1}{*}{$\surd$}} &
      \multicolumn{1}{c}{\multirow{1}{*}{\textbf{97.9 $\pm$ 0.2}}} &
      \multicolumn{1}{c|}{\multirow{1}{*}{\textbf{4.1}}} &
      \multicolumn{1}{c}{\multirow{1}{*}{\textbf{97.8 $\pm$ 0.3}}} &
      \multicolumn{1}{c}{\multirow{1}{*}{\textbf{4.6}}} \\

       \Xhline{1pt}
       
  \end{tabular} 
   \label{Tab:Ablation_Visual}
\end{table}

\vspace{1mm}
\noindent
\textbf{Comparison on Zero-Shot Tasks.}
In \cref{Fig:Comparison}, we compare the generalization performance of different methods on zero-shot tasks. The results show that our model achieves substantially higher success rates and shorter task completion times than the competing models across multiple zero-shot tasks. These results demonstrate that incorporating spatiotemporal representations into both the visual and action modalities enables our model to maintain strong generalization performance, even on unseen robotic tasks.

\vspace{1mm}
\noindent
\textbf{Comparison on SpatioTemporal Planning.}
In \cref{Fig:Comparison_Spatiotemporal}, we compare the predicted spatiotemporal action trajectories produced by representative 2D, 3D, and 4D models: OpenVLA \cite{kim2024openvla}, SpatialVLA \cite{qu2025spatialvla} and our VLA-4D. The results show that the 2D model exhibits substantial redundant global motion and pronounced oscillations in local motion speed. The 3D model generates much smoother global trajectories, but the local motion speed still fluctuates noticeably. In contrast, the 4D model produces both smooth global trajectories and stable local motion speeds. These results demonstrate that our VLA-4D provides a new and reliable paradigm for achieving spatiotemporally coherent robotic manipulation.

\begin{table}\footnotesize
    \setlength{\abovecaptionskip}{3pt}
    \setlength\tabcolsep{1.5pt}
    \setlength{\belowcaptionskip}{3pt}
    \caption{Effect of action representation components.} 
  \centering
  \renewcommand\arraystretch{1.3}
  \begin{tabular}{ccccc}
    \Xhline{1pt}
      \multicolumn{1}{c|}{\multirow{2}{*}{Action representation}} & \multicolumn{2}{c|}{\multirow{1}{*}{LIBERO-Spatial}} & \multicolumn{2}{c}{\multirow{1}{*}{LIBERO-Goal}} \\

      \cline{2-5}
      \multicolumn{1}{c|}{\multirow{1}{*}{}} &
      \multicolumn{1}{c}{\multirow{1}{*}{Succ(\%)$\uparrow$}} &
      \multicolumn{1}{c|}{\multirow{1}{*}{Time(s)$\downarrow$}} &
      \multicolumn{1}{c}{\multirow{1}{*}{Succ(\%)$\uparrow$}} &
      \multicolumn{1}{c}{\multirow{1}{*}{Time(s)$\downarrow$}} \\

      \hline 

      \multicolumn{1}{c|}{\multirow{1}{*}{Spatial param.}} &
      \multicolumn{1}{c}{\multirow{1}{*}{96.8 $\pm$ 0.3}} &
      \multicolumn{1}{c|}{\multirow{1}{*}{5.0}} &
      \multicolumn{1}{c}{\multirow{1}{*}{97.1 $\pm$ 0.3}} &
      \multicolumn{1}{c}{\multirow{1}{*}{5.7}} \\

      \multicolumn{1}{c|}{\multirow{1}{*}{Spatial + Temporal param.}} &
      \multicolumn{1}{c}{\multirow{1}{*}{\textbf{97.9 $\pm$ 0.2}}} &
      \multicolumn{1}{c|}{\multirow{1}{*}{\textbf{4.1}}} &
      \multicolumn{1}{c}{\multirow{1}{*}{\textbf{97.8 $\pm$ 0.3}}} &
      \multicolumn{1}{c}{\multirow{1}{*}{\textbf{4.6}}} \\

       \Xhline{1pt}
       
  \end{tabular} 
   \label{Tab:Ablation_Action}
\end{table}

\subsection{Ablation Study}
\label{sec:abalation}
\noindent
\textbf{Effect of Visual Representation Modules.}
In \cref{Tab:Ablation_Visual}, we validate the effectiveness of spatial embedding, temporal embedding, and feature fusion modules in the visual representation. The results show that both the spatial and temporal embeddings are key to significantly improving the overall performance of our model by a large margin. Furthermore, the feature fusion further enhances the upper limit of the performance of our model on robotic manipulation tasks.

\vspace{1mm}
\noindent
\textbf{Effect of Action Representation Components.}
In \cref{Tab:Ablation_Action}, we compare the effects of spatial and temporal components in the action representation. When only the spatial action representation is introduced, our model already achieves a relatively high task success rate, but with a higher task completion time. After further 
\gh{incorporation of} temporal representation, the success rate improves slightly, 
but the task completion time decreases significantly. These results indicate that the spatial parameters determine the effectiveness of action planning, 
\gh{and} the temporal parameters further enhance its efficiency and superiority.

\begin{table}\footnotesize
    \setlength{\abovecaptionskip}{3pt}
    \setlength\tabcolsep{2pt}
    \setlength{\belowcaptionskip}{3pt}
    \caption{Impact of various input modalities on robotic manipulation.} 
  \centering
  \renewcommand\arraystretch{1.3}
  \begin{tabular}{ccccccc}
    \Xhline{1pt}
      \multicolumn{1}{c}{\multirow{2}{*}{\makecell{Vision\\data}}} & 
      \multicolumn{1}{c}{\multirow{2}{*}{4D cues}} &
      \multicolumn{1}{c|}{\multirow{2}{*}{Proprio.}} &
      \multicolumn{2}{c|}{\multirow{1}{*}{LIBERO-Spatial}} & \multicolumn{2}{c}{\multirow{1}{*}{LIBERO-Goal}} \\

      \cline{4-7}
      \multicolumn{1}{c}{\multirow{1}{*}{}} &
      \multicolumn{1}{c}{\multirow{1}{*}{}} &
      \multicolumn{1}{c|}{\multirow{1}{*}{}} &
      \multicolumn{1}{c}{\multirow{1}{*}{Succ(\%)$\uparrow$}} &
      \multicolumn{1}{c|}{\multirow{1}{*}{Time(s)$\downarrow$}} &
      \multicolumn{1}{c}{\multirow{1}{*}{Succ(\%)$\uparrow$}} &
      \multicolumn{1}{c}{\multirow{1}{*}{Time(s)$\downarrow$}} \\

      \hline 

      \multicolumn{1}{c}{\multirow{1}{*}{Image}} &
      \multicolumn{1}{c}{\multirow{1}{*}{$\times$}} &
      \multicolumn{1}{c|}{\multirow{1}{*}{$\times$}} &
      \multicolumn{1}{c}{\multirow{1}{*}{85.9 $\pm$ 0.6}} &
      \multicolumn{1}{c|}{\multirow{1}{*}{5.9}} &
      \multicolumn{1}{c}{\multirow{1}{*}{88.0 $\pm$ 0.8}} &
      \multicolumn{1}{c}{\multirow{1}{*}{6.5}} \\

      \multicolumn{1}{c}{\multirow{1}{*}{Video}} &
      \multicolumn{1}{c}{\multirow{1}{*}{$\times$}} &
      \multicolumn{1}{c|}{\multirow{1}{*}{$\times$}} &
      \multicolumn{1}{c}{\multirow{1}{*}{89.2 $\pm$ 0.6}} &
      \multicolumn{1}{c|}{\multirow{1}{*}{5.7}} &
      \multicolumn{1}{c}{\multirow{1}{*}{90.1 $\pm$ 0.7}} &
      \multicolumn{1}{c}{\multirow{1}{*}{6.3}} \\

      \multicolumn{1}{c}{\multirow{1}{*}{Video}} &
      \multicolumn{1}{c}{\multirow{1}{*}{$\surd$}} &
      \multicolumn{1}{c|}{\multirow{1}{*}{$\times$}} &
      \multicolumn{1}{c}{\multirow{1}{*}{97.1 $\pm$ 0.2}} &
      \multicolumn{1}{c|}{\multirow{1}{*}{4.1}} &
      \multicolumn{1}{c}{\multirow{1}{*}{97.3 $\pm$ 0.4}} &
      \multicolumn{1}{c}{\multirow{1}{*}{4.6}} \\

      \multicolumn{1}{c}{\multirow{1}{*}{Video}} &
      \multicolumn{1}{c}{\multirow{1}{*}{$\surd$}} &
      \multicolumn{1}{c|}{\multirow{1}{*}{$\surd$}} &
      \multicolumn{1}{c}{\multirow{1}{*}{\textbf{97.9 $\pm$ 0.2}}} &
      \multicolumn{1}{c|}{\multirow{1}{*}{\textbf{4.1}}} &
      \multicolumn{1}{c}{\multirow{1}{*}{\textbf{97.8 $\pm$ 0.3}}} &
      \multicolumn{1}{c}{\multirow{1}{*}{\textbf{4.6}}} \\

       \Xhline{1pt}
       
  \end{tabular} 
   \label{Tab:Ablation_Input}
\end{table}

\begin{figure}
  \setlength{\abovecaptionskip}{7pt}
  \setlength{\belowcaptionskip}{-5pt}
  \centering
   \includegraphics[width=0.99\linewidth]{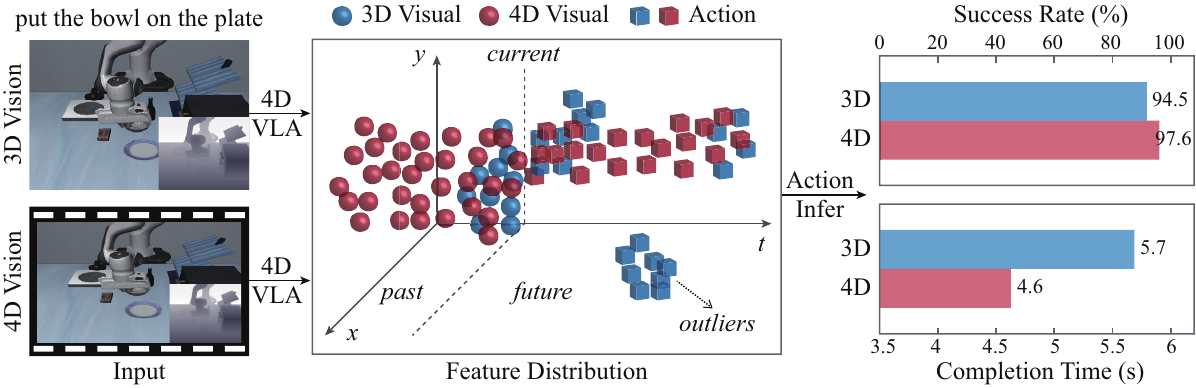}
   \caption{Influence of visual representations on actions. Compared with 3D features, 4D visual features allow the VLA to exploit past visual cues, producing accurate and coherent actions.
   }
   \label{Fig:Discussion_Mechanism}
\end{figure}

\vspace{1mm}
\noindent
\textbf{Influence of Input Modality.}
In \cref{Tab:Ablation_Input}, we conduct an ablation study on the impacts of different input modalities: vision data (\emph{e.g.}, images and videos), 4D cues, and proprioceptive states. When only image-based vision input is used, our model achieves a moderate task success rate. When switching to video-based vision inputs, the success rate improves slightly. With the further introduction of 4D cues, both the task success rate and \gh{the} completion time improve significantly. When incorporating proprioceptive states, our model yields a minor but consistent overall performance gain. These results suggest that video-based vision data ensure a preliminary baseline capability, 4D cues substantially enhance model performance, and proprioceptive states provide an additional complementary improvement.

\subsection{Discussion}
\label{sec:discussion}
\noindent
\textbf{How 4D Vision Affects Spatiotemporal Action?}
In \cref{Fig:Discussion_Mechanism}, we analyze the interaction mechanism between 4D-aware visual representations and spatiotemporal action representations through feature distribution visualization. Specifically, we compare the 3D visual representations (derived from image+depth input) and their corresponding action representations with the 4D visual representations (derived from video+depth input) and their corresponding action representations under the same task settings. First, the 3D visual feature distribution lacks the temporal dimension. As a result, the corresponding action features are spatially clustered but temporally scattered. This leads to robotic operations that achieve a reasonable success rate but incur \gh{a} longer completion time.
Second, the 4D visual feature distribution forms a continuous spatiotemporal manifold, and the corresponding action features are spatiotemporally clustered and continuous.
Consequently, the robot achieves the highest task success rate with a reduced completion time. Within our VLA-4D framework, 
spatial knowledge embedded in 4D visual features enhances 
spatial localization of robotic actions, \gh{and} 
temporal knowledge improves their temporal efficiency during manipulation.

\begin{table}\footnotesize
    \setlength{\abovecaptionskip}{3pt}
    \setlength\tabcolsep{1.4pt}
    \setlength{\belowcaptionskip}{3pt}
    \caption{Discussion on spatiotemporal encoding.} 
  \centering
  \renewcommand\arraystretch{1.3}
  \begin{tabular}{ccccc}
    \Xhline{1pt}
      \multicolumn{1}{c|}{\multirow{2}{*}{4D cues}} & \multicolumn{2}{c|}{\multirow{1}{*}{LIBERO-Spatial}} & \multicolumn{2}{c}{\multirow{1}{*}{LIBERO-Goal}} \\

      \cline{2-5}
      \multicolumn{1}{c|}{\multirow{1}{*}{}} &
      \multicolumn{1}{c}{\multirow{1}{*}{Succ(\%)$\uparrow$}} &
      \multicolumn{1}{c|}{\multirow{1}{*}{Time(s)$\downarrow$}} &
      \multicolumn{1}{c}{\multirow{1}{*}{Succ(\%)$\uparrow$}} &
      \multicolumn{1}{c}{\multirow{1}{*}{Time(s)$\downarrow$}} \\

      \hline 

      \multicolumn{1}{c|}{\multirow{1}{*}{w/o Spatiotemporal encoding}} &
      \multicolumn{1}{c}{\multirow{1}{*}{96.3 $\pm$ 0.4}} &
      \multicolumn{1}{c|}{\multirow{1}{*}{4.3}} &
      \multicolumn{1}{c}{\multirow{1}{*}{96.1 $\pm$ 0.3}} &
      \multicolumn{1}{c}{\multirow{1}{*}{4.9}} \\

      \multicolumn{1}{c|}{\multirow{1}{*}{w/ Spatiotemporal encoding}} &
      \multicolumn{1}{c}{\multirow{1}{*}{\textbf{97.9 $\pm$ 0.2}}} &
      \multicolumn{1}{c|}{\multirow{1}{*}{\textbf{4.1}}} &
      \multicolumn{1}{c}{\multirow{1}{*}{\textbf{97.8 $\pm$ 0.3}}} &
      \multicolumn{1}{c}{\multirow{1}{*}{\textbf{4.6}}} \\

       \Xhline{1pt}
       
  \end{tabular} 
   \label{Tab:Discussion_STEncoding}
\end{table}

\begin{table}\footnotesize
    \setlength{\abovecaptionskip}{3pt}
    \setlength\tabcolsep{3pt}
    \setlength{\belowcaptionskip}{3pt}
    \caption{Impact of various visual feature fusion strategies.} 
  \centering
  \renewcommand\arraystretch{1.3}
  \begin{tabular}{ccccc}
    \Xhline{1pt}
      \multicolumn{1}{c|}{\multirow{2}{*}{Fusion strategy}} & \multicolumn{2}{c|}{\multirow{1}{*}{LIBERO-Spatial}} & \multicolumn{2}{c}{\multirow{1}{*}{LIBERO-Goal}} \\

      \cline{2-5}
      \multicolumn{1}{c|}{\multirow{1}{*}{}} &
      \multicolumn{1}{c}{\multirow{1}{*}{Succ(\%)$\uparrow$}} &
      \multicolumn{1}{c|}{\multirow{1}{*}{Time(s)$\downarrow$}} &
      \multicolumn{1}{c}{\multirow{1}{*}{Succ(\%)$\uparrow$}} &
      \multicolumn{1}{c}{\multirow{1}{*}{Time(s)$\downarrow$}} \\

      \hline 

      \multicolumn{1}{c|}{\multirow{1}{*}{Concatenation}} &
      \multicolumn{1}{c}{\multirow{1}{*}{94.2 $\pm$ 0.4}} &
      \multicolumn{1}{c|}{\multirow{1}{*}{4.5}} &
      \multicolumn{1}{c}{\multirow{1}{*}{94.9 $\pm$ 0.4}} &
      \multicolumn{1}{c}{\multirow{1}{*}{4.8}} \\

      \multicolumn{1}{c|}{\multirow{1}{*}{Weighting}} &
      \multicolumn{1}{c}{\multirow{1}{*}{95.8 $\pm$ 0.3}} &
      \multicolumn{1}{c|}{\multirow{1}{*}{4.3}} &
      \multicolumn{1}{c}{\multirow{1}{*}{96.5 $\pm$ 0.3}} &
      \multicolumn{1}{c}{\multirow{1}{*}{4.8}} \\

      \multicolumn{1}{c|}{\multirow{1}{*}{Attention}} &
      \multicolumn{1}{c}{\multirow{1}{*}{\textbf{97.9 $\pm$ 0.2}}} &
      \multicolumn{1}{c|}{\multirow{1}{*}{\textbf{4.1}}} &
      \multicolumn{1}{c}{\multirow{1}{*}{\textbf{97.8 $\pm$ 0.3}}} &
      \multicolumn{1}{c}{\multirow{1}{*}{\textbf{4.6}}} \\

       \Xhline{1pt}
       
  \end{tabular} 
   \label{Tab:Discussion_Fusion}
\end{table}

\begin{table}\footnotesize
    \setlength{\abovecaptionskip}{3pt}
    \setlength\tabcolsep{3pt}
    \setlength{\belowcaptionskip}{3pt}
    \caption{Comparison between different training strategies.} 
  \centering
  \renewcommand\arraystretch{1.3}
  \begin{tabular}{ccccc}
    \Xhline{1pt}
      \multicolumn{1}{c|}{\multirow{2}{*}{Training strategy}} & \multicolumn{2}{c|}{\multirow{1}{*}{LIBERO-Spatial}} & \multicolumn{2}{c}{\multirow{1}{*}{LIBERO-Goal}} \\

      \cline{2-5}
      \multicolumn{1}{c|}{\multirow{1}{*}{}} &
      \multicolumn{1}{c}{\multirow{1}{*}{Succ(\%)$\uparrow$}} &
      \multicolumn{1}{c|}{\multirow{1}{*}{Time(s)$\downarrow$}} &
      \multicolumn{1}{c}{\multirow{1}{*}{Succ(\%)$\uparrow$}} &
      \multicolumn{1}{c}{\multirow{1}{*}{Time(s)$\downarrow$}} \\

      \hline 

      \multicolumn{1}{c|}{\multirow{1}{*}{Only Stage 2}} &
      \multicolumn{1}{c}{\multirow{1}{*}{91.2 $\pm$ 0.5}} &
      \multicolumn{1}{c|}{\multirow{1}{*}{4.9}} &
      \multicolumn{1}{c}{\multirow{1}{*}{90.7 $\pm$ 0.6}} &
      \multicolumn{1}{c}{\multirow{1}{*}{5.3}} \\

      \multicolumn{1}{c|}{\multirow{1}{*}{Stage 1 + Stage 2}} &
      \multicolumn{1}{c}{\multirow{1}{*}{\textbf{97.9 $\pm$ 0.2}}} &
      \multicolumn{1}{c|}{\multirow{1}{*}{\textbf{4.1}}} &
      \multicolumn{1}{c}{\multirow{1}{*}{\textbf{97.8 $\pm$ 0.3}}} &
      \multicolumn{1}{c}{\multirow{1}{*}{\textbf{4.6}}} \\

       \Xhline{1pt}
       
  \end{tabular} 
   \label{Tab:Discussion_Training}
\end{table}

\vspace{1mm}
\noindent
\textbf{Role of Spatiotemporal Encoding.}
We study the importance of spatiotemporal encoding on positions and timestamps for our model. As demonstrated in \cref{Tab:Discussion_STEncoding}, the model equipped with spatiotemporal encoding achieves better performance on robotic manipulation tasks compared to the one without it. This is because spatial positions and temporal information operate on different scales, which 
\gh{can} lead to gradient conflicts during training. In contrast, a dedicated encoding strategy can map 
spatiotemporal information into a unified feature space 
\gh{to ensure} stable convergence and \gh{prevention of feature space collapse.} 

\vspace{1mm}
\noindent
\textbf{Choice of Visual Feature Fusion Strategy.}
\cref{Tab:Discussion_Fusion} compares the impact of different visual feature fusion strategies: concatenation, weighting, and attention. The results show that attention-based fusion outperforms other fusion strategies. This is because concatenation and weighting rely on global unified fusion with fixed weights. In contrast, attention-based fusion can dynamically adjust the fusion weights of visual features according to 4D spatiotemporal embeddings. This allows our model to focus more on meaningful 4D-aware visual features for action prediction.

\vspace{1mm}
\noindent
\textbf{Impact of Training Strategy.}
In \cref{Tab:Discussion_Training}, we study the effects of different training strategies on our model. Compared to using only stage 2 for fine-tuning, the multi-stage training strategy that combines stage 1 and stage 2 leads to a substantial improvement in model performance on robotic manipulation tasks. The underlying reason is that stage 1 training enhances the spatiotemporal visual reasoning ability through large-scale 3D and 4D vision-language pretraining, facilitating more accurate spatiotemporal action prediction for coherent robotic planning.

\vspace{1mm}
\noindent
\textbf{Limitation.}
Our VLA-4D can predict fine-grained action parameters for spatiotemporally coherent robotic manipulation, but still faces challenges when deployed in unseen real-world environments. The main reason is that robotic operations in unseen environments may introduce action errors due to uncontrollable factors such as mechanical wear and calibration drift, which in turn reduce the overall efficiency of manipulation. In the future, we plan to incorporate reinforcement learning \cite{sutton1998reinforcement} to correct the errors online in predicted spatiotemporal actions, enabling more robust and adaptive robotic planning in real-world environments.

\section{Conclusion}
\label{sec:conclusion}
In this work, we propose VLA-4D, a general 4D vision-language-action model for spatiotemporally coherent robotic manipulation.
We design a 4D-aware visual representation that fuses 3D positions and 1D time into visual features for fine-grained spatiotemporal reasoning. Moreover, we construct a spatiotemporal action representation that incorporates temporal variables into spatial parameters for fine-grained spatiotemporal planning. By performing multimodal alignment, our model can produce precise and coherent actions for the robot. To support training, we extend the representative VLA dataset with temporal action annotations to fine-tune our model for spatiotemporal robotic manipulation. Extensive experiments \gh{and ablations on benchmark datasets} verify the superiority of our method.

{
  \small
  \bibliographystyle{ieeenat_fullname}
  \bibliography{egbib}
}

\end{document}